\documentclass{article}

\RequirePackage{amsmath}
\RequirePackage[noblocks]{authblk}
\RequirePackage{hyperref} 
\RequirePackage{amssymb}
\RequirePackage{graphicx}

\usepackage{makecell}
\usepackage[table]{xcolor}
\usepackage[linesnumbered]{algorithm2e}
\makeatletter

\SetAlgorithmName{Pseudocode}{}{}

\let\oldnl\nl

\newcommand{\nonl}{\renewcommand{\nl}{\let\nl\oldnl}}
\usepackage{comment}
\usepackage{amssymb}
\usepackage{pifont}
\usepackage{wrapfig}
\usepackage{capt-of}
\usepackage{arydshln}
\usepackage{dirtytalk}
\usepackage{xcolor}
\usepackage{tablefootnote}
\newcommand{\cmmnt}[1]{\ignorespaces}  
\usepackage{lineno}
\makeatletter
\def\makeLineNumberLeft{%
  \linenumberfont\llap{\hb@xt@\linenumberwidth{\LineNumber\hss}\hskip\linenumbersep}
  \hskip\columnwidth
  \rlap{\hskip\linenumbersep\hb@xt@\linenumberwidth{\hss\LineNumber}}\hss}
\leftlinenumbers
\makeatother

\newcommand{\beginsupplement}{%
        \setcounter{table}{0}
        \renewcommand{\thetable}{S\arabic{table}}%
        \setcounter{figure}{0}
        \renewcommand{\thefigure}{S\arabic{figure}}%
        \setcounter{algocf}{0}
        \renewcommand{\thealgocf}{S\arabic{algocf}}%
     }

\graphicspath{ {./Figures/}}
\usepackage{framed}
\usepackage{subcaption}
\usepackage{float}

\RequirePackage{fullpage}
\RequirePackage{fancyhdr}

\newcommand{\HRule}{\rule{\linewidth}{0.5mm}}

\makeatletter
\renewcommand{\@maketitle}{%
  \parindent=0pt
  \centering
  {\Large \bfseries\textsc{\@title}}
  \HRule\par%
  \textit{\@author \hfill}
  \par
}

\makeatletter
\def\thickhline{%
  \noalign{\ifnum0=`}\fi\hrule \@height \thickarrayrulewidth \futurelet
   \reserved@a\@xthickhline}
\def\@xthickhline{\ifx\reserved@a\thickhline
               \vskip\doublerulesep
               \vskip-\thickarrayrulewidth
             \fi
      \ifnum0=`{\fi}}
\makeatother

\newlength{\thickarrayrulewidth}
\usepackage{xspace}
\usepackage{xcolor}

\newcommand{\chgd}[1]{\textcolor{violet}{#1}}

\usepackage[round]{natbib}
\usepackage{multirow}
\makeatother 

\providecommand{\keywords}[1]{\textbf{\textit{Keywords:}} #1}

\usepackage[none]{hyphenat}

\usepackage{array}  

\begin{document}
\sloppy

\title{SOMTimeS: Self Organizing Maps for Time Series Clustering and its Application to Serious Illness Conversations} 

\author[*,1]{Ali Javed} 
\author[2]{Donna M. Rizzo}
\author[1]{Byung Suk Lee}
\author[3]{Robert Gramling}

\affil[1]{Department of Computer Science, University of Vermont, Burlington, VT, USA}
\affil[2]{Department of Civil \& Environmental Engineering, University of Vermont, Burlington, VT, USA}
\affil[3]{Department of Family Medicine, University of Vermont, Burlington, VT, USA}

\affil[*]{Corresponding author email: ali.javed@uvm.edu}
\affil[*]{Mailing Address: 82 University Place, Innovation STEM Building, Burlington, VT 05405, United States}

\maketitle 

\bigskip
    

\begin{abstract}

There is an increasing demand for scalable algorithms capable of clustering and analyzing large time series datasets. The Kohonen self-organizing map (SOM) is a type of unsupervised artificial neural network for visualizing and clustering complex data, reducing the dimensionality of data, and selecting influential features. Like all clustering methods, the SOM requires a measure of similarity between input data (in this work time series). Dynamic time warping (DTW) is one such measure, and a top performer given that it accommodates the distortions when aligning time series. Despite its use in clustering, DTW is limited in practice because it is quadratic in runtime complexity with the length of the time series data. To address this, we present a new DTW-based clustering method, called SOMTimeS (a Self-Organizing Map for TIME Series), that scales better and runs faster than other DTW-based clustering algorithms, and has similar performance accuracy. The computational performance of SOMTimeS stems from its ability to prune unnecessary DTW computations during the SOM’s training phase. We also implemented a similar pruning strategy for K-means for comparison with one of the top performing clustering algorithms. We evaluated the pruning effectiveness, accuracy, execution time and scalability on 112 benchmark time series datasets from the University of California, Riverside classification archive. We showed that for similar accuracy, the speed-up achieved for SOMTimeS and K-means was 1.8x on average; however, rates varied between 1x and 18x depending on the dataset. SOMTimeS and K-means pruned 43\% and 50\% of the total DTW computations, respectively. We applied SOMtimeS to natural language conversation data collected as part of a large healthcare cohort study of patient-clinician serious illness conversations to demonstrate the algorithm's utility with complex, temporally sequenced phenomena.

\end{abstract}




\section{Introduction}\label{SOMsec:Introduction}

By 2025, it is estimated that more than four hundred and fifty exabytes of data will be collected and stored daily~\citep{WEF2019}. Much of that data will be collected continuously and represent phenomena that change over time. We propose that fully understanding the meaning of these data will often require complexity scientists to model them as time series. Examples include data collected by sensors~\citep{IoT1,Cisco1}, every day natural language~\citep[e.g.,][]{Bentley2018,ROSS2020826,Reagan2016,Chu2017}, biomonitors~\citep{Gharehbaghi2018}, waterflow, barometric pressure and other routine environmental condition meters~\citep[e.g.,][]{Hamami2020,JAVED2020125802,EWEN2011178}, social media interactions~\citep[e.g.,][]{Bie2016,JAVED201823,Javed2016,Javed_SpringerCCIS}, and hourly financial data reported by fluctuating world stock and currency markets~\citep{Lasfer2013}. In response to the increasing amounts of time-oriented data available to analysts, the applications of time-series modeling are growing rapidly~\citep[e.g.,][]{Minaudo2017,Racmi2015,MATHER2015,Bende-Michl2013,Iorio2018,Gupta2018,Pirim2012,deSouto2008,Flangan2017}.

Time series modeling is computationally ``expensive'' in terms of processing power and speed of analysis. Indeed, as the numbers of observations or measurement dimensions for each observation increase, the relative efficiency of time series modeling diminishes, creating an exponential deterioration in computational speed. Under conditions where computing power is in excess or when the speed for generating results is not of concern, these challenges would be less pressing. However, these conditions are rarely met currently, and the accelerating rate of data collection promises to continue outpacing the computational infrastructure available to most analysts. 

In this work, we embed distance-pruning into a K-means and a new artificial neural network  -- \textbf{S}elf-\textbf{O}rganizning \textbf{M}ap for \textbf{t}ime \textbf{s}eries (SOMTimeS) to improve the execution time of clustering methods that used Dynamic Time Warping (DTW) for large time series applications. The computational efficiency of these algorithms is attributed to the pruning of unnecessary DTW computations during the training phases of each algorithm. When assessed using 112 time series datasets from the University of California, Riverside (UCR) classification archive, SOMTimeS and K-means prunded 43\% and 50\% of the DTW computations, respectively. On average there is a 2x speed-up when clustering all 112 of the archived datasets. The pruning efficiency and resulting speed-up vary depending on the dataset being clustered. However, to the best of our knowledge, K-means with DTW distance pruning and SOMTimeS are the fastest DTW-based clustering algorithms to date.

SOMTimeS is designed to leverage the outstanding visualization capabilities of the SOM when clustering problems of high complexity. To explore the potential utility of SOMTimeS in this regard, to the natural sequential ordering of data, we evaluated its performance when applied to the science of doctor-family-patient conversations in high emotion settings. Understanding and improving serious illness communication is a national priority for 21st century healthcare, but our existing methods for measuring and analyzing such data is cumbersome, human intensive, and far too slow to be relevant for large epidemiological studies, communication training or time-sensitive reporting. Here, we use data from an existing multi-site epidemiological study of healthcare serious illness conversations as one example of how efficient computational methods can add to the science of healthcare communication.

The remainder of this paper is organized as follows. 
Section~\ref{SOMsec:preliminaries} provides background information on SOMs and DTW. 
Section~\ref{SOMsec:SOMTimeS} presents the SOMTimeS algorithm.
Section~\ref{SOMsec:evaluation} and Section~\ref{SOMsec:Case_Study} evaluate the results of SOMTimeS as well as two DTW-based clustering algorithms -- K-means and TADPole on the UCR benchmark datasets and serious illness discussions, respectively.
Section~\ref{SOMsec:Discussions} discusses the results. 
Section~\ref{SOMsec:Conclusion} concludes the paper and suggests future work. 

\section{Background}\label{SOMsec:preliminaries}

Similar to the work of \cite{Silva2020}, \cite{Li202}, \cite{Parshutin2008} and, \cite{Somervuo1999}, SOMTimeS is a new artificial neural network that embeds distance-pruning strategy into a DTW-based Kohonen Self-Organizing Map. While the Kohonen SOM (see details in Section~\ref{SOMsec:SOM}) is linearly scalable with respect to the number of input data, it often performs hundreds of passes (i.e., epochs) when self-organizing or clustering the training data. Each epoch requires $n \times M$ distance calculations, where $n$ is the number of observations and $M$ is the number of nodes in the network map. This large number of distance calculations is problematic, particularly when the distance measure is computationally expensive, as is the case with DTW (see Section~\ref{SOMsec:dtw}). 

DTW, originally introduced in 1970s for speech recognition~\citep{Sakoe1978}, continues to be one of the more robust, top performing, and consistently chosen learning algorithms for time series data~\citep{Xiaopeng2006,Ding2008,Paparrizos2016,Paparrizos2017,Nurjahan2016,JAVED2020100001}. Its ability to shift, stretch, and squeeze portions of the time series helps address challenges inherent to time series data (e.g., optimize the alignment of two temporal sequences). Unfortunately, the ability to align the temporal dimension comes with increased computational overhead, which has hindered its use in practical applications involving large datasets or long time series clustering~\citep{JAVED2020100001,Zhu2012}. The first subquadratic-time algorithm ($O(m^2/\log\log{m})$) for DTW computation was proposed by~\cite{Gold2018}, which is still more computationally expensive in comparison to the simpler Euclidean distance ($O(m)$).

To address the computational cost, several studies have presented approximate solutions\cmmnt{ to improve the feasibility of DTW computation in clustering}~\citep{Zhu2012,Salvador2007,Ghazi2009}. To the best of our knowledge, TADPole by \cite{Nurjahan2016} is the only algorithm (see supplementary material Section~\ref{SOMsec:TADPole}) that speeds up the DTW computation without using an approximation. It does so by using a bounding mechanism to prune the expensive DTW calculations. Yet, when coupled with the clustering algorithm (i.e., Density Peaks of~\cite{Rodriguez1492}), it still scales quadratically. Thus, even after decades of research~\citep{Zhu2012,Nurjahan2016,Lou2015,Slavador2007,Renjie2020}, the almost quadratic time complexity of DTW-based clustering still poses a challenge when clustering time series in practice.

\subsection{Self Organizing Maps}\label{SOMsec:SOM}
\begin{figure}[!ht]
\centering
\includegraphics[width=1\textwidth]{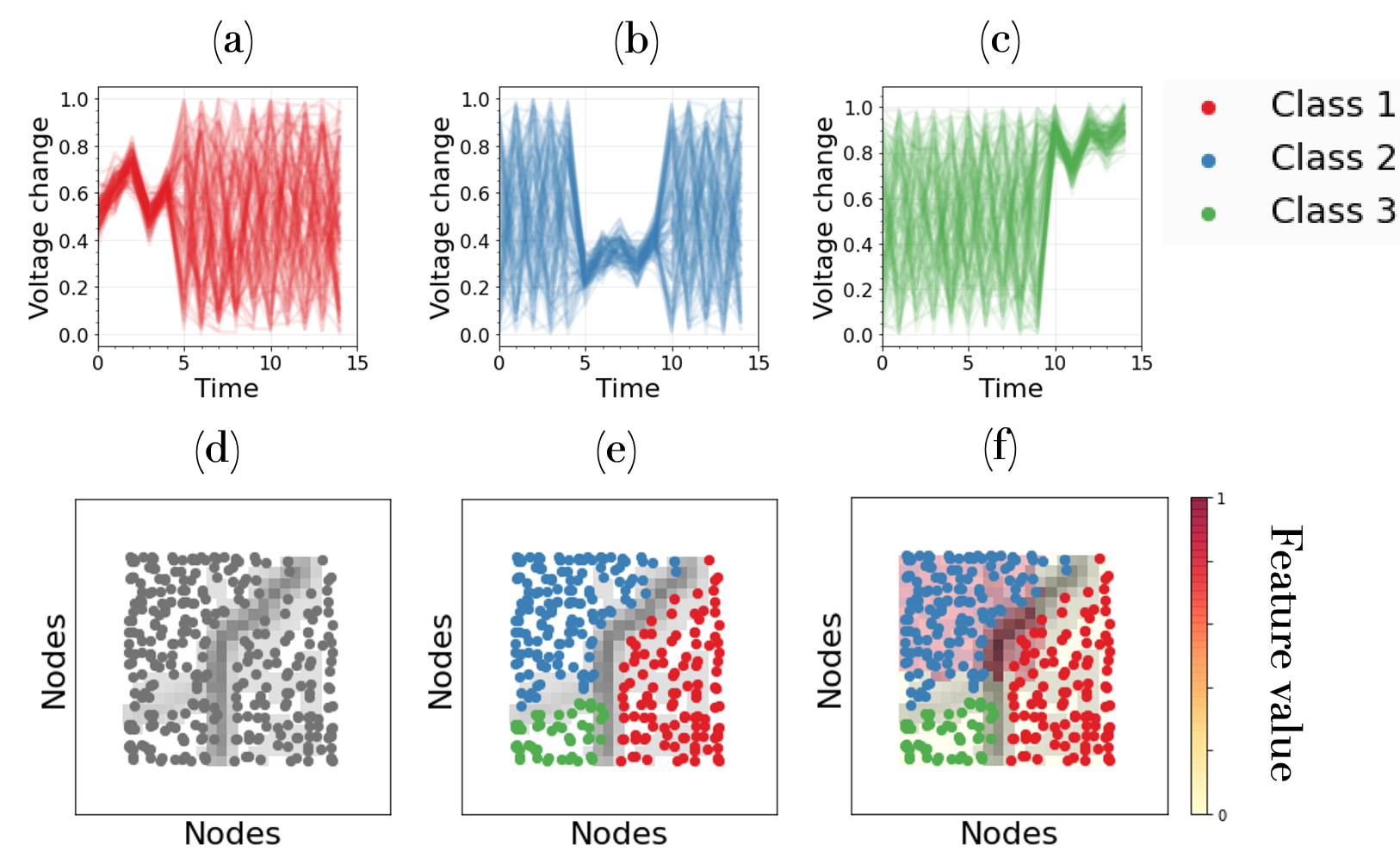}
\caption{Clustering and visualizing times series observations from one of the UCR archive datasets --- InsectEPGRegularTrain, using a Self-Organized Map with DTW as the distance measure. Panels (a) -- (c) show the three classified time series associated with voltage changes that captures interactions between insects and their food source (e.g., plants). Panel (d) shows clustered observations superimposed on a unified distance matrix, (e) color-coded clusters, and (f) a single input variable (or feature value) in the background.}\label{SOMfig:SOM}
\end{figure}
The Kohonen Self-Organizing Map~\citep{Kohonen2001,KOHONEN2013} may be used to either cluster or classify observations, and has advantages when visualizing complex, nonlinear data~\citep{ALVAREZGUERRA2008,Eshghi2011}. Additionally, it has been shown to outperform other parametric methods on datasets containing outliers or high variance~\citep{MANGIAMELI1996}. Similar to methods such as logistic regression and principal component analysis, SOMs may be used for feature selection, as well as mapping input data from a high-dimensional space to a lower-dimensional space (typically a two-dimensional mesh or lattice). The DTW-based SOM clusters data from one of the UCR archive datasets (InsectEPGRegularTrain) onto a 2-D mesh (see Figure~\ref{SOMfig:SOM}). The dataset represents voltage changes of an electrical circuit that captures the interaction between insects and their food source (e.g., plants). These data had already been classified into three categories (see Figures~\ref{SOMfig:SOM}a, b, and c).
Each gray dot for Figure~\ref{SOMfig:SOM}d represents a time series (i.e., temporal pattern or arc).  The self-organized observations may be plotted with what is known as a unified distance matrix or U-matrix~\citep{Ultsch1993}. The latter is obtained by calculating the average difference between the weights of adjacent nodes in the trained SOM, and then plotting these values (in a gray scale of Figure~\ref{SOMfig:SOM}(d)) on the trained 2-D mesh. Darker shading represents higher U-matrix values (larger average distance between observations). In this manner, the U-matrix can help assess the quality and the number of clusters. For example, see the U-matrix of Figure~\ref{SOMfig:SOM}(e), which separates the observations into three clusters that may be color-coded or labeled (should labels exist). Finally, any information, input features, or metadata associated with the observations may be visualized or superimposed (red shading of Figure~\ref{SOMfig:SOM}(f)) in the same 2-D space in order to explore associations and the importance of individual input features with the clustered results. The ability to visualize individual input features in the same space as the clustered observations (known as component planes) makes the SOM a powerful tool for data analysis and feature selection.

\subsection{Dynamic time warping}\label{SOMsec:dtw}
\begin{figure}[!ht]
\centering
\includegraphics[width=0.8\textwidth]{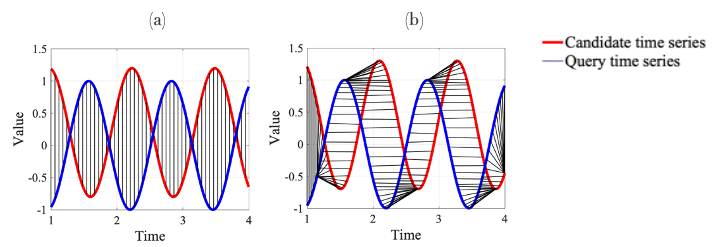}
\caption{Alignment between two times series for calculating (a) Euclidean distance and (b) DTW distance.}\label{SOMfig:DTW_EUC_SOM}
\end{figure}

DTW is recognized as one of the most accurate similarity measures for time series data~\citep{Paparrizos2017, Rakthanmanon2012, johnpaul2020}. While the most common measure, Euclidean distance, uses a one-to-one alignment between two time series (e.g., labeled candidate and query in Figure~\ref{SOMfig:DTW_EUC_SOM}(a)), DTW employs a one-to-many alignment that warps the time dimension (see Figure~\ref{SOMfig:DTW_EUC_SOM}(b)) in order to minimize the sum of distances between time series samples. As such, DTW can optimize alignment both globally (by shifting the entire time series left or right) and locally (by stretching or squeezing portions of the time series). The optimal alignment should adhere to three rules: 
\begin{enumerate}
    \item Each point in the query time series must be aligned with one or more points from candidate time series, and vice versa.
    \item The first and last points of the query and a candidate time series must align with each other. 
    \item No cross-alignment is allowed; that is, the aligned time series indices must increase monotonically.
\end{enumerate}

DTW is often restricted to aligning points only within a moving window of a fixed size to improve accuracy and reduce computational cost. The window size may be optimized using supervised learning on the training data. When supervised learning is not possible (i.e., clustering), a window size amounting to 10\% of the observation data is usually considered adequate~\citep{chotirat2004}.

\subsubsection{Upper and lower bounds for DTW-based distance metric}\label{SOMsec:dtw_bound}
SOMTimeS uses distance bounding to prune the DTW calculations performed during the SOM unsupervised learning. This distance bounding involves finding a tight upper and lower bound. Because DTW is designed to find a mapping that minimizes the sum of the point-to-point distances between two time series, that mapping can never result in a summed distance that is greater than the sum of point-to-point Euclidean distance. Hence, finding the tight upper bound is straight forward -- it is the Euclidean distance~\citep{Keogh2002}. 
To find the lower bound, we use a method -- the LB\_Keogh method~\citep{Keogh2003}, common in similarity searches~\citep{Keogh2003,Ratanamahatana2005,Wei2005} and clustering~\citep{Nurjahan2016}.
\begin{figure}[!ht]
\centering
\includegraphics[width=0.8\textwidth]{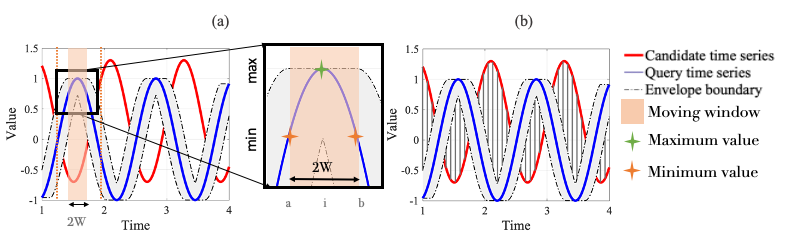}
\caption{Two steps of calculating the L\_Keogh tight lower bound for DTW in linear time: (a) determine the envelope around a query time series, and (b) sum the point to point distance shown in grey lines between the envelope and a candidate time series as LB\_Keogh (Equation\,\ref{eq:lb_keogh}).}\label{SOMfig:LB_Keogh}
\end{figure} 
The LB\_Keogh method comprises two steps (see Figure~\ref{SOMfig:LB_Keogh}a and Figure~\ref{SOMfig:LB_Keogh}b). Given a fixed DTW window size, $W$, one of the two time series (called the query time series, Q) is bounded by an envelope having an upper ($U_i$) and lower boundary ($L_i$) calculated at time step $i$, respectively, as:
\begin{equation}
\begin{aligned}
U_i = \max(q_{a},...,q_{i},...,q_{b}) \\
L_i = \min(q_{a},...,q_{i},...,q_{b}),  \label{eq:L_U}
\end{aligned}
\end{equation}\label{eq:LBKeogh}
\noindent where $a = i-W$, and $b= i+W$ (see Figure~\ref{SOMfig:LB_Keogh}a).
In the second step, the LB\_Keogh lower bound is calculated as the sum of Euclidean distance between the candidate time series and the envelope boundaries (see vertical lines of Figure~\ref{SOMfig:LB_Keogh}b). Equation~\ref{eq:lb_keogh} shows the formula for calculating the LB\_Keogh lower bound:
\begin{equation}
\text{LB\_Keogh}= \sqrt{ \sum_{i=1}^{m}{
\begin{cases} (t_i - U_i)^2,& \text{if } t_i > U_i\\
(t_i - L_i)^2, & \text{if } t_i < L_i\\
0, & \text{otherwise}
\end{cases} }} \label{eq:lb_keogh}
\end{equation}
\\
where $t_i$, $U_i$, and $L_i$ are the values of a candidate time series, the upper and lower envelope boundary, respectively, at time step $i$.

\section{The SOMTimeS Algorithm}\label{SOMsec:SOMTimeS}

\begin{figure}[!ht]
\centering
\includegraphics[width=0.6\textwidth]{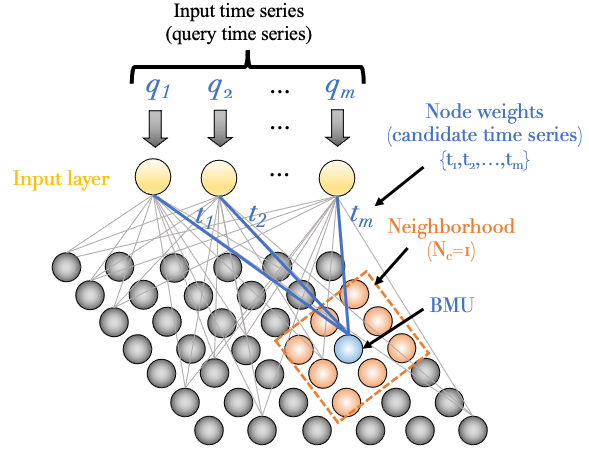}
\caption{Schematic of the Kohonen Self-Organizing Map (after Kohnen, 2001) showing weights (candidate time series) of the best matching unit (BMU) in blue surrounded by a user-specified neighborhood ($N_c$).}\label{SOMfig:SOM_input}
\end{figure}

SOMTimeS is a variant of the SOM (see Pseudocode~\ref{alg:SOMTimeS}), where each input observation (i.e., query time series) is compared with the weights (i.e., candidate time series) associated with each node in the 2-D SOM mesh (see Figure~\ref{SOMfig:SOM_input}). During training, the comparison (or distance calculation) between these two time series is performed to identify the SOM node whose weights are most similar to a given input time series; this node is identified as the ``best matching unit (BMU)''. Once the nodal weights (candidate time series) of the BMU have been identified, these weights (and those of the neighborhood nodes) are updated to more closely match the query time series (Line~\ref{lin:update} of Pseudocode~\ref{alg:SOMTimeS}). This same process is performed for all query time series in the dataset -- defined as one epoch. While iterating through some user-defined fixed number of epochs, both the neighborhood size and the magnitude of change to nodal weights are incrementally reduced. This allows the SOM to converge to a solution (stable map of clustered nodes), where the set of weights associated with these self-organized nodes now approximate the input time series (i.e., observed data). In SOMTimeS, the distance calculation is done using DTW with bounding, which helps prune the number of DTW calculations required to identify the BMU.
\begin{figure}[!ht]
\centering
\includegraphics[width=0.65\textwidth]{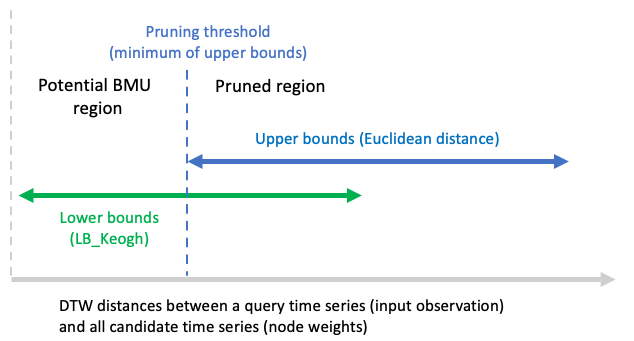}
\caption{Identification of a qualification region in SOMTimeS.}\label{SOMfig:bounding}
\end{figure}

\subsection{Pruning of DTW computations}
Pruning is performed in two steps. 
First, an upper bound (i.e., Euclidean distance) is calculated between the input observation and each weight vector associated with the SOM nodes (Line~\ref{lin:upperbound} of Pseudocode~\ref{alg:SOMTimeS}). The minimum of these upper bounds is set as the pruning threshold (see dotted line in Figure~\ref{SOMfig:bounding}). Next, for each SOM node we calculate a lower bound (i.e., LB\_Keogh; see Line~\ref{lin:lowerbound}). If the calculated lower bound is greater than the pruning threshold, that respective node is pruned from being the BMU. If the lower bound is less than the pruning threshold, then that SOM node lies in what we refer to as the potential BMU region (see Figure~\ref{SOMfig:bounding}, and Line~\ref{lin:dtw}). As a result, the more expensive DTW calculations are performed only for the nodes in this potential BMU region. The node having the minimum summed distance is the BMU.

After identifying the BMUs for each input time series, the BMU weights, as well as the weights of nodes in some neighborhood of the BMUs, are updated to more closely match the respective input time series using a traditional learning algorithm based on gradient descent (Line~\ref{lin:update} of Pseudocode~\ref{alg:SOMTimeS}). Both the learning rate and the neighborhood size are reduced (see lines~\ref{lin:parameterUpdate1} and~\ref{lin:parameterUpdate2}) over each epoch until the nodes have self-organized (i.e., algorithm has converged).  In this work, unless otherwise stated, SOMTimeS is trained for 100 epochs. To further reduce the SOM execution time, the set of input (i.e., query) time series  may be partitioned in a manner similar to~\cite{WU1991821},~\cite{OBERMAYER1990381} and ~\cite{Lawrence1999}  for parallel processing (see Line~\ref{lin:split}).
We should also note that after convergence, SOMTimeS may be used to \emph{classify} observations into a given number of clusters should a known number of classes exist. This is done by setting the mesh size equal to $k$ (i.e., desired number of classes), and using the weights of the BMUs for direct class assignment.

\renewcommand\baselinestretch{1}\selectfont
\begin{algorithm}
\caption{SOMTimeS algorithm pseudocode}\label{alg:SOMTimeS} 
\nonl \underline{Algorithm SOMTimeS pseudocode}\\
\SetAlgoLined
\nonl Input: a set $\mathcal{S}$ of query time series $\{Q_1, Q_2, ..., Q_n\}$, \textit{epochs}: number of \textit{epochs}, $W$: warping window size\\
\nonl Assumption: Similarity between two observation is the DTW distance between their time series.\\
\nonl Output: \text{Best Matching Units}
\smallskip\\
\nonl Procedure
\smallskip\\
mesh\_size $M$ := $5 \times \sqrt{n}$\qquad\qquad // $n$ = the number of time series in $\mathcal{S}$\\
Create and randomly initialize a mesh of Nodes[$\sqrt{M}$,$\sqrt{M}$], where the weights of each node are a randomly-generated time series (candidate time series) of length equal to the query time series.\\
neighborhood size $N_c$ := $\sqrt{M}/2$\\
learning rate $r$ := 0.9 \\
Split $\mathcal{S}$ into subsets of equal size, $\mathcal{S}_1, \mathcal{S}_2, ..., \mathcal{S}_c$, where $c$ is the number of available CPU cores in the machine.\\ \label{lin:split}

\For {each epoch $p$}{
    \For {each split $\mathcal{S}_i$ ($i=1,2,...,c$) assigned to the core $i$ in parallel}
        {
        \For {each input time series $Q_{i_j} (j=1,2,...,n/c)$ in $\mathcal{S}_i$} {
            upper bounds:= Euclidean distances between $Q_{i_j}$ and weights of each node.\label{lin:upperbound}\\  %
            lower bounds:= $LB\_Keogh$ between $Q_{i_j}$ and weights of each node using the $W$. \label{lin:lowerbound} \\
        \nonl // Prune the set of all nodes to the set of qualified nodes.\\ 
        $Qualified$:= Set of nodes whose weights have a lower bound with $Q_{i_j}$  $\leq$ $\min(\text{upper bounds})$. \label{lin:qualifiedTS} \\
        Best matching unit:= Compute DTW distance between $Q_{i_j}$ and weights of nodes in $Qualified$. The best matching unit (BMU) is the node whose weights are most similar to $Q_{i_j}$.\label{lin:dtw}
    }
}

\For {each time series $Q_i (i = 1, 2, ..., n)$}
    {
         \nonl // Update the node weights of the BMU (and its neighborhood) identified for $Q_i$ using a gradient descent based on learning rate, to more closely match $Q_i$.\\ 
         
        \For{Weights ($t_{1..m}$) of BMU and its neighbor nodes}{
        $t_m$ = $t_m + r \times (q_m - t_m)$ \label{lin:update}
        }
        
    }
\nonl {\tt // Update the neighborhood size and the learning rate in SOM.}\\ 
$N_c$ :=  $\sqrt{M}/2 \times (1 - p / \textit{\#epochs})$ \label{lin:parameterUpdate1}\\
$r$ := $0.9 \times (1 - p / \textit{\#epochs})$ \label{lin:parameterUpdate2}\\
}
return Best matching units
\nonl \end{algorithm}
\renewcommand\baselinestretch{2}\selectfont


\section{Performance Evaluations}\label{SOMsec:evaluation}

The UCR time series classification archive~\citep{UCRArchive2018}, with thousands of citations and downloads, is arguably the most popular archive for benchmarking time series clustering algorithms. The archive was born out of frustration, with studies on clustering and classification reporting error rates on a single time series dataset, and then implying that the results would generalize to other datasets. At the time of this writing, the archive has 128 datasets comprising a variety of synthetic, real, raw and pre-processed time series data, and has been used extensively for benchmarking the performance of clustering algorithms~\citep[e.g.,][]{Paparrizos2016,Paparrizos2017,Nurjahan2016,JAVED2020100001,Zhu2012}. To evaluate SOMTimeS, we excluded sixteen of the archive datasets because they contained only a single cluster, or had time series lengths that vary. The latter prohibited a fair comparison of SOMTimeS to DTW-based K-means. The remaining 112 datasets were used to evaluate the accuracy, execution time, and scalability of SOMTimeS. We fixed the DTW window constraint at 5\% of the length of the observation data following earlier recommendations by~\cite{Paparrizos2016,Paparrizos2017}.

\subsection{Algorithm Assessment}

Accuracy is reported using six assessment metrics (see Table~\ref{SOMtbl:accuracy_time} and Table~\ref{SOMtbl:accuracy_time_2}) that include the Adjusted Rand Index (ARI)~\citep{Santos2009}, Adjusted Mutual Information (AMI)~\citep{Romano2016}, the Rand Index (RI)~\citep{hubert1985}, Homogeneity~\citep{Rosenberg2017}, Completeness~\citep{Rosenberg2017}, and Fowlkes Mallows index (FMS)~\citep{Fowlkes1983}.
Scalability and execution time of DTW-based clustering algorithms are inversely affected by the length and total number of times series being clustered or classified. As a result, we report the number of DTW computations and execution time as a function of \emph{problem size}, defined as $\sum_{i=1}^{n}{|Q|_i}$, where $|Q|$ is the length of times series $Q$, and $n$ is the total number of time series in the dataset. The presence of a few large datasets in the archive makes it more informative to visualize problem size as the \chgd{logarithm to the base 10} (see Figure~\ref{SOMfig:problemSize} in Supplementary Material). Finally, for comparison purposes, the same assessment metrics are reported for the three more popular clustering algorithms that use DTW as a distance measure --- 1) K-means, 2) the SOM, and 3) TADPole. We embedded the same pruning strategy into K-means for a more equitable comparison of speed-up and clustering quality.

\subsubsection{Pruning speed-up}\label{SOMsec:accuracy}
The performance of the DTW-based SOM and K-means may be quantified in two important ways: 1) execution speed and 2) clustering quality using six different assessment indices. When pruning is not used and the number of passes (i.e., epochs) through the dataset are fixed at 10, the SOM and K-means require 13 and 14 hours, respectively, to cluster all 112 datasets in the UCR archive with comparable assessment indices (see Table~\ref{SOMtbl:accuracy_time}).
The SOM, however, typically requires more passes through the dataset than K-means to achieve optimal performance. When 100 epochs are used the SOM achieves a higher accuracy than K-means for 5 of the 6 measures (see Table~\ref{SOMtbl:accuracy_time}); however, the execution time increases almost linearly to $\sim148$ hours.
\begin{table}[!htb]
\begin{small}
\caption{Assessment of execution times and clustering quality averaged over the 112 datasets in the UCR archive for K-means and the SOM clustering methods (without pruning). Note: The six assessment indices (usually expressed as values between $0$ and $1$) have been multiplied by $100$; metric averages closer to 100 represent better performance.}\label{SOMtbl:accuracy_time}
\begin{tabular}{|c|c|c|c|c|c|c|c|c|c|c|c|c|c|}
\hline
\multirow{2}{*}{\thead{\bf Algorithm \\ (without pruning)}} & \multirow{2}{*}{\bf Hours*} &\multicolumn{2}{|c|}{\bf ARI \tablefootnote{Adjusted Rand Index~\citep{Santos2009}}}  & \multicolumn{2}{|c|}{\bf AMI \tablefootnote{Adjusted Mutual Information~\citep{Romano2016}}}&\multicolumn{2}{|c|}{\bf RI \tablefootnote{Rand Index~\citep{hubert1985}}}&\multicolumn{2}{|c|}{\bf H\tablefootnote{Homogeneity~\citep{Rosenberg2017}}}&\multicolumn{2}{|c|}{\bf C\tablefootnote{Completeness~\citep{Rosenberg2017}}}&\multicolumn{2}{|c|}{\bf FMS\tablefootnote{Fowlkes Mallows index~\citep{Fowlkes1983}}}\\
\cline{3-14}
& &avg & std & avg & std & avg & std & avg & std & avg & std & avg & std\\
\hline
\hline
\thead{\bf K-means - DTW \\ \bf 10-iterations}  & 13 & 22 & 23 & 28 & 26 & 66 & 22 & 29 & 26 &  40 & 32 & 49 & 19\\
\hline
\thead{\bf SOM - DTW \\ 10-Epochs}  & \textbf{14} & \textbf{21} & \textbf{23} & \textbf{27} & \textbf{24} & \textbf{70} & \textbf{15} & \textbf{28} & \textbf{24} &  \textbf{31} & \textbf{26} & \textbf{47} & \textbf{20}\\
\hline
\thead{\bf SOM - DTW \\ \bf 100-epochs} & 148 & 24 & 23 & 30 & 26 & 71 & 16 & 31 & 25 &  35 & 28 & 50 & 19\\
\hline
\multicolumn{14}{l}{%
  \begin{minipage}{15cm}%
  \vspace{0.5em}
    *All algorithms were executed on same computational machine --- dual 20-Core Intel Xeon E5-2698 v4 2.2 GHz machine with 512 GB 2,133 MHz DDR4 RDIMM.%
  \end{minipage}%
}
\end{tabular}
\end{small}
\end{table}

\begin{figure}[!htb]
\centering
\begin{subfigure}[t]{0.4\textwidth}
\centering
\caption{SOMTimeS}\label{SOMfig:speedup_SOMTimeS}
\includegraphics[width=1\textwidth]{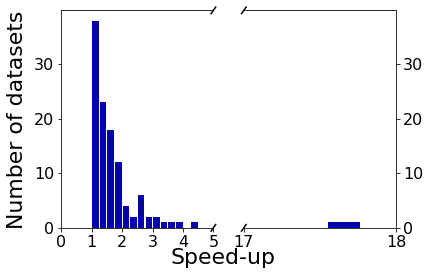}
\end{subfigure}
\begin{subfigure}[t]{0.4\textwidth}
\centering
\caption{K-means}\label{SOMfig:speedup_Kmeans}
\includegraphics[width=1\textwidth]{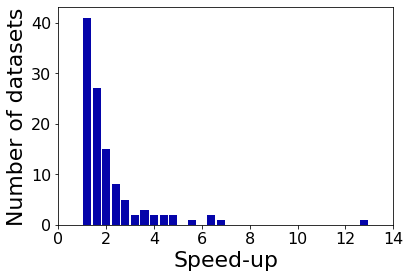}
\end{subfigure}
\caption{Speed-up factor achieved for different datasets in the UCR archive.}\label{SOMfig:speedup}
\end{figure}

Table~\ref{SOMtbl:accuracy_time_2} summarizes the execution time and six assessment indices when both the DTW-based K-means and SOM (i.e., SOMTimeS) clustering algorithms use the pruning mechanism. Additionally we present the results of TADPole~\citep{Nurjahan2016}, since it also uses DTW distance pruning to speedup clustering. 
While the speed-up times vary by dataset, DTW distance pruning improves the execution time by a factor of 2x (on average) when clustering all the 112 datasets in the UCR archive (see Figure~\ref{SOMfig:speedup}). 
The clustering results with or without pruning are identical, and therefore, the quality (assessment indices) is the same. In comparison to TADPole, SOMTimeS is 17x faster and has a higher value in 5 out of 6 assessment indices.
\begin{table}[!htb]
\begin{small}
\caption{Assessment of execution times and clustering quality averaged over the 112 datasets in the UCRarchive for 1) DTW-based K-means, 2) SOMTimeS and 3) TADPole with pruning. Note: Assessment indices (usually expressed as values between $0$ and $1$) have been multiplied by $100$; metric averages closer to 100 represent better performance.}\label{SOMtbl:accuracy_time_2}
\begin{tabular}{|c|c|c|c|c|c|c|c|c|c|c|c|c|c|}
\hline
\multirow{2}{*}{\thead{\bf Algorithm \\ (with pruning)}} & \multirow{2}{*}{\bf Hours*} &\multicolumn{2}{|c|}{\bf ARI$^1$}  & \multicolumn{2}{|c|}{\bf AMI$^2$}&\multicolumn{2}{|c|}{\bf RI$^3$}&\multicolumn{2}{|c|}{\bf H$^4$}&\multicolumn{2}{|c|}{\bf C$^5$}&\multicolumn{2}{|c|}{\bf FMS$^6$}\\
\cline{3-14}
& &avg & std & avg & std & avg & std & avg & std & avg & std & avg & std\\
\hline
\hline
\thead{\bf K-means \\ 10-iterations}  & \textbf{6} & \textbf{22} & \textbf{23} & \textbf{28} & \textbf{26} & \textbf{66} & \textbf{22} & \textbf{29} & \textbf{26} &  \textbf{40} & \textbf{32} & \textbf{49} & \textbf{19}\\
\thead{\bf SOMTimeS \\ 10-Epochs}  & \textbf{7} & \textbf{21} & \textbf{23} & \textbf{27} & \textbf{24} & \textbf{70} & \textbf{15} & \textbf{28} & \textbf{24} &  \textbf{31} & \textbf{26} & \textbf{47} & \textbf{20}\\
\thead{\bf SOMTimeS \\ 100-epochs}  & \textbf{58} & \textbf{24} & \textbf{23} & \textbf{30} & \textbf{26} & \textbf{71} & \textbf{16} & \textbf{31} & \textbf{25} &  \textbf{35} & \textbf{28} & \textbf{50} & \textbf{19}\\
\textbf{TADPole}  & \textbf{1011} & \textbf{16} & \textbf{25} & \textbf{24} & \textbf{27} & \textbf{62} & \textbf{18} & \textbf{25} & \textbf{26} &  \textbf{36} & \textbf{31} & \textbf{51} & \textbf{20}\\
\hline
\multicolumn{14}{l}{%
  \begin{minipage}{15cm}%
  \vspace{0.5em}
    *All algorithms were executed on same computational machine --- dual 20-Core Intel Xeon E5-2698 v4 2.2 GHz machine with 512 GB 2,133 MHz DDR4 RDIMM.%
  \end{minipage}%
}
\end{tabular}
\end{small}
\end{table}

When SOMTimeS is executed in parallel, the reduction in execution time is a factor of the number of CPUs available. To cluster all 112 datasets in the UCR archive, SOMTimeS (at 100 epochs) took 3 hours using 20 CPUs, and took only 20 minutes when the number of SOM epochs was set to 10. Similar parallization can be achieved for K-means, and TADPole.

\begin{figure}[!htb]
\centering
\includegraphics[width=0.8\textwidth]{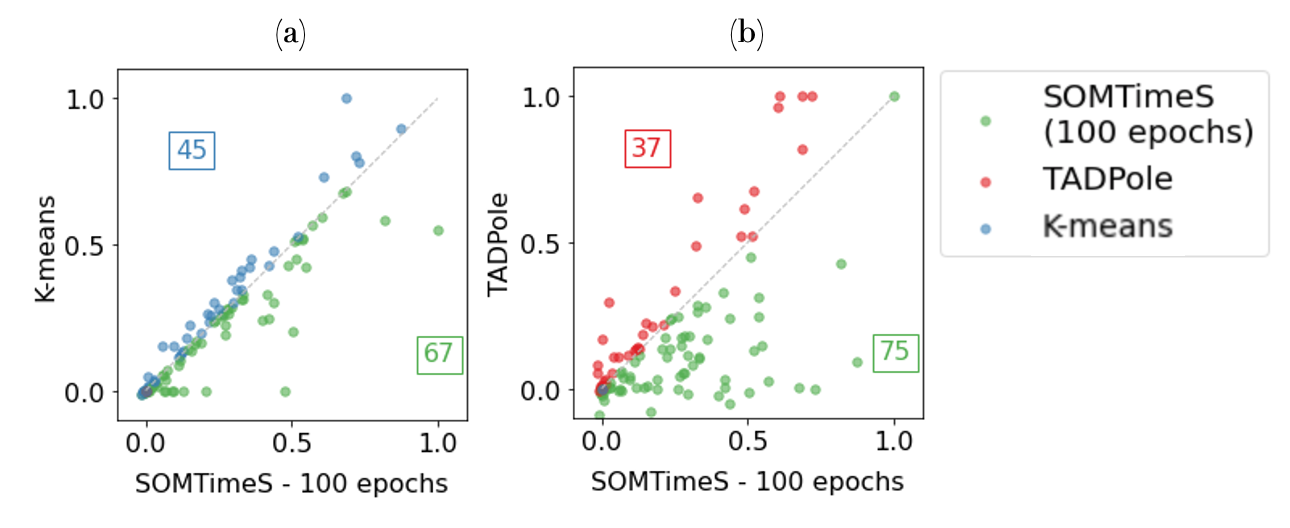}
\caption{ARI scores for SOMTimeS (shown in green) vs. (a) TADPole (red), and (b) K-means (blue) across all 112 of the UCR datasets.}\label{SOMfig:accuracy}
\label{SOMfig:measures}
\end{figure}
Because the ARI (Adjusted Rand Index) is recommended as one of the more robust measures for assessing accuracy across datasets~\citep{Milligan1986ASO,JAVED2020100001}, we plot the ARI scores for SOMTimeS (at 100 epochs) against K-means, and TADPole (Figures~\ref{SOMfig:accuracy} (a) and (b), respectively) for each of the 112 URC datasets. The same clustering algorithm has high variation in performance metrics across datasets~\citep{JAVED2020100001}. The green points (67 of the 112 datasets) lying below the 45-degree line of panel (a) represent higher accuracy for SOMTimeS, while the ARI scores above the diagonal (shown in blue) indicate that K-means outperforms SOMTimeS for 45 of the 112 datasets. 
The comparison of ARI scores for SOMTimeS and TADPole (Figure~\ref{SOMfig:accuracy}b) shows higher accuracy for SOMTimeS for 75 of the 112 datasets, and lower accuracy for the remaining 37 datasets.

\subsubsection{Execution time and scalability}
%
As mentioned previously, the speed-up achieved for SOMTimeS and K-means is a result of the pruning strategy. We study the effects of the pruning in four ways --- 1) percentage of DTW computations pruned as function of time series length, 2) the total number of DTW computations pruned, 3) the scalability as a function of DTW computations performed, and 4) the change in the rate of DTW pruning over epochs.

\begin{figure}[!htb]
\centering
\includegraphics[width=0.8\textwidth]{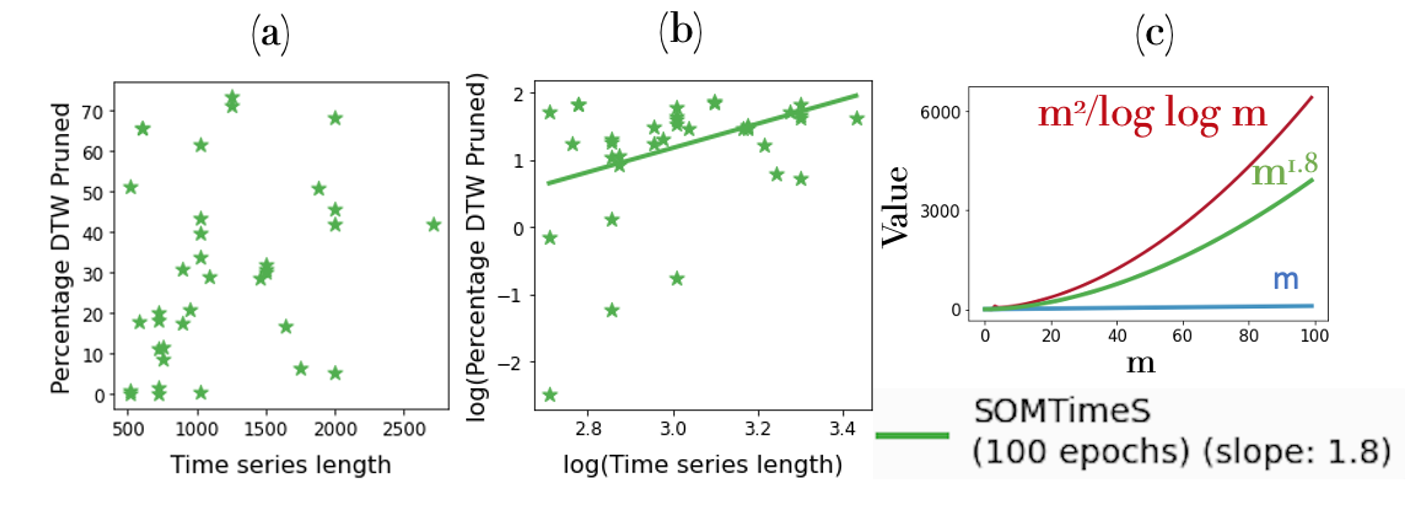}
\caption{Percentage of DTW computations pruned 
with respect to the time series length shown on (a) 
linear scale, and (b) log-log scale axes; green stars represents one of the 36 UCR archived datasets, and (c) empirical approximation of the pruning rate as a function of time series length (m).}\label{SOMfig:sequence_length}
\end{figure}

\textbf{Percentage of pruning with respect to the length of individual time series}: Because DTW scales with the length of individual time series, we examined the number of DTW computations pruned as a function of time series length. Figure~\ref{SOMfig:sequence_length} shows the percentage of DTW computations pruned 
for increasing time series length on both linear and log-log scaled axes. Figure~\ref{SOMfig:sequence_length}a shows a subset (n=36) of the UCR archived datasets. Here the subset comprises all datasets where the total number of time series is greater than 100 and the length of time series is greater than 500. Figure~\ref{SOMfig:sequence_length}b shows the corresponding log-log plot, where the slope approximates the relationship between pruning rate and time series length. This increase in pruning rate is close to the DTW complexity of ($(m^2/\log\log{m})$), where $m$ is the length of time series (see Figure~\ref{SOMfig:sequence_length}c).

\textbf{Total number of pruned DTW computations:} 
K-means pruned more than 50\% of the DTW calculations for 34 of the datasets, where as SOMTimeS (with epochs set to 10 and 100, respectively) pruned more than 50\% of the DTW calculations for 8 and 21 of the 112 UCR datasets, respectively. TADPole pruned more than 50\% of the DTW calculations for 40 of the datasets (see Figure~\ref{SOMfig:pruning_10_100} in Supplementary Material). Despite the apparent pruning advantage of TADPole, its quadratic complexity O($n^2$) with respect to DTW calculations (compared to O($n$) in SOMTimeS) results in more DTW computations, particularly for larger datasets.

\begin{figure}[!htb]
\centering
\includegraphics[width=0.8\textwidth]{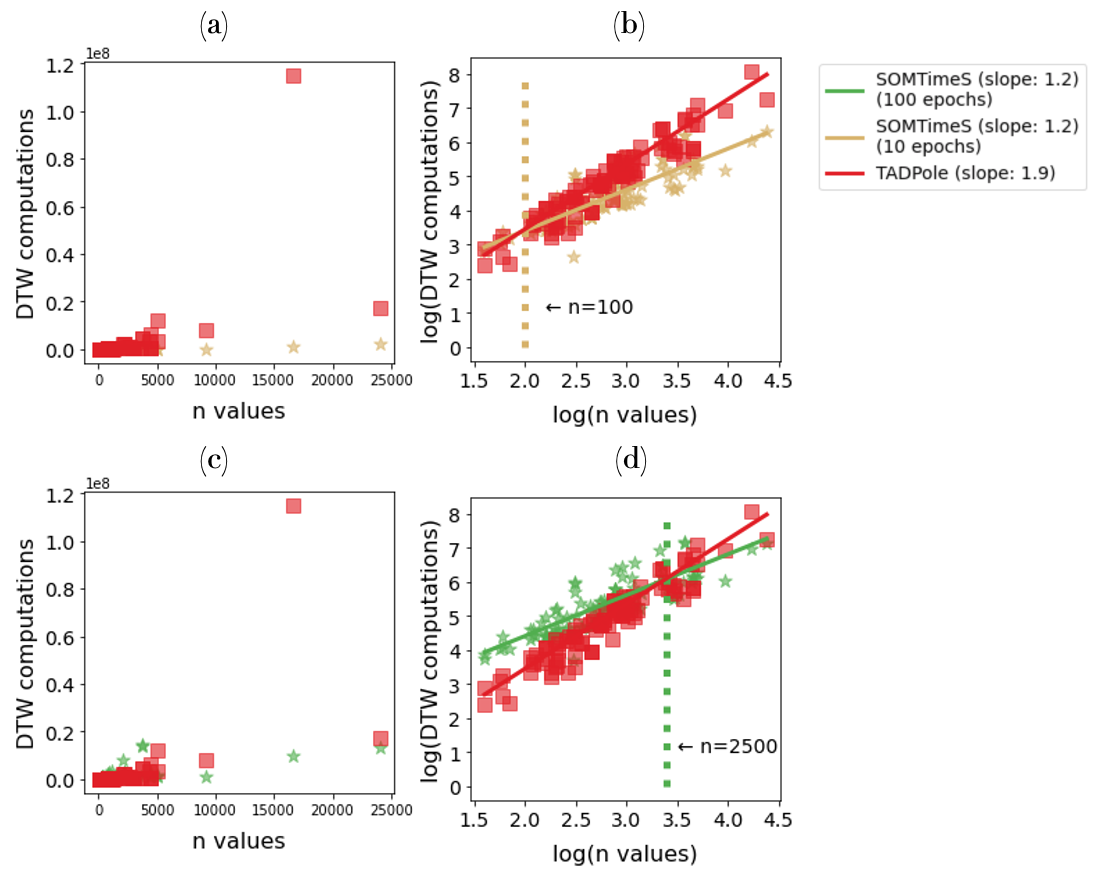}
\caption{DTW computations performed as a function of dataset size shown on linear scaled axes (panels a and c) and log-log scale (panels b and d) for TADPole (200 million computations total) and SOMTimeS (13 million computations total at 10 epochs, and 100 million computations at 100 epochs).}\label{SOMfig:CallsMade}
\end{figure}

\textbf{Scaling of DTW computations performed as a function of number of input time series:} Because TADPole performs O($n^2$) DTW calculations, the number of calls to DTW increases quadratically with the number $n$ of input time series. The threshold (in terms of the number of input time series, $n$) after which the number of calls to the DTW function in SOMTimeS is less than that of TADPole, depends on the number of epochs. This cutoff is empirically observed to be close to $n = 100$ and $n = 2500$, for 10 and 100 epochs, respectively (see Figure~\ref{SOMfig:CallsMade}). SOMTimeS and K-means have similar theoretical and empirical complexities when the mesh size in SOMTimeS is set equal to $k$.

Overall, when clustering over all 112 of the UCR archived datasets, SOMTimeS computed the DTW measure 13 million and 100 million times (at 10 and 100 epochs, respectively). K-means computed the DTW measure 8 millions times; while TADPole by comparison computed DTW 200 million times (see Figure~\ref{SOMfig:CallsMade}). At a dataset level, SOMTimeS had fewer calls for 12 of the datasets (when using 10 epochs) compared to K-means. At 100 epochs, SOMTimeS requires more calculations than K-means at 10 iterations to cluster all 112 datasets. In comparison to TADPole, SOMTimeS had fewer calls for 88 of the datasets (when using 10 epochs), and 26 of the datasets (for 100 epochs). However, the quality of clustering for SOMTimeS at 100 epochs increases for 4 of the six assessment indices compared to K-means, and TADPole.

\begin{figure}[!htb]
\centering

\begin{subfigure}[b]{0.4\textwidth}
\centering
\caption{}\label{SOMfig:dtwOverEpoch}
\includegraphics[width=1\textwidth]{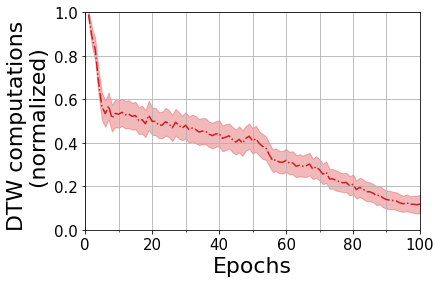}
\end{subfigure}
\quad%
\begin{subfigure}[b]{0.4\textwidth}
\centering
\caption{}\label{SOMfig:timeOverEpoch}
\includegraphics[width=1\textwidth]{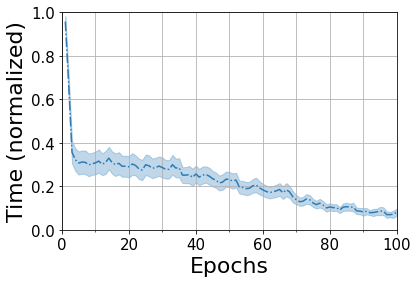}
\end{subfigure}
\caption{Change in the SOMTimeS pruning 
effect as the number of epochs increases  
measured as the normalized (a) number of calls to the DTW function
and (b) execution time.  The 
dashed line represents the mean value for all datasets after individually normalizing each dataset run over all epochs. The shaded region corresponds to 95\% confidence interval around the mean.}
\label{SOMfig:overEpochs}
\end{figure}

\textbf{Change in the SOMTimeS pruning rate as a function of epochs:} When we examine the pruning 
effect as a function of epochs, both the number of DTW calls and the execution time decrease as the number of epochs increases. As the nodes of the SOM mesh organize, more nodes get pruned; and hence, fewer nodes exist in the unpruned region (i.e., potential BMU region of Figure~\ref{SOMfig:bounding}), which decreases the need for DTW calls. 
Figure~\ref{SOMfig:dtwOverEpoch} shows the total number of calls to the DTW function 
made for each dataset, normalized over all 
epochs. The dashed line represents the  average number of calls over all datasets and the shaded region shows the 95\% confidence interval. Figure~\ref{SOMfig:timeOverEpoch} shows the corresponding normalized execution time. Both normalized DTW calls and execution time per epoch steadily decrease with increasing number of epochs and iterative updating of SOM weights. The elbow point, where further 
epochs result 
in a diminishing reduction of DTW calculations,
is at 
the 6th epoch. This is called the swapover point and occurs when the self-organizing map moves from gross reorganization of the SOM weights to fine-tuning of the weights. 

\begin{figure}[!htb]
\centering
\includegraphics[width=0.8\textwidth]{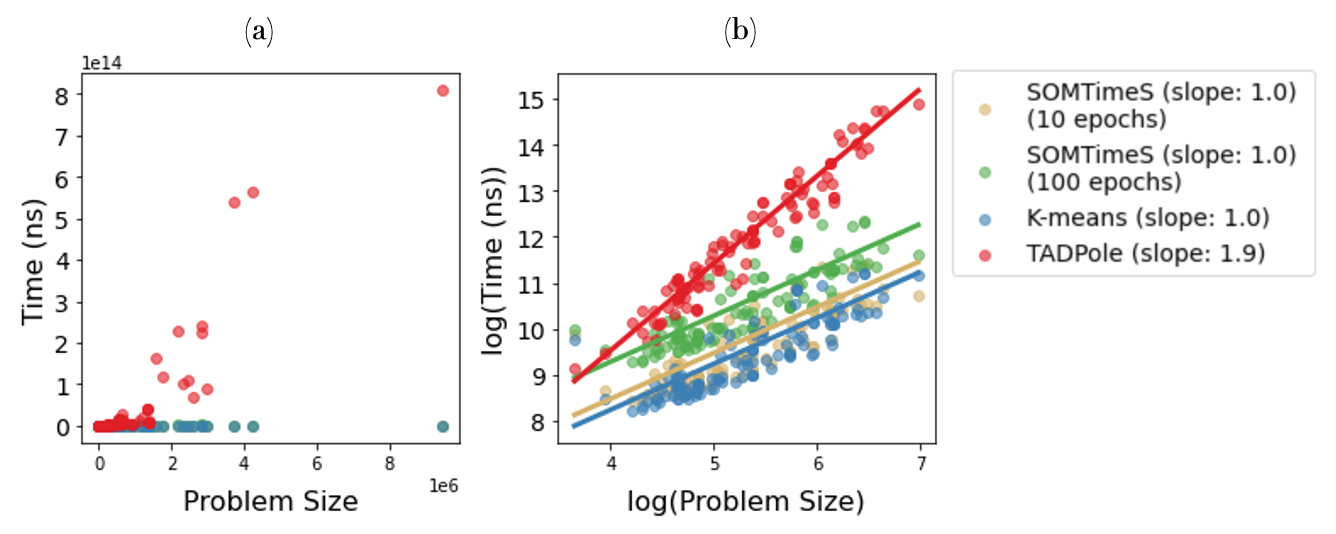}
\caption{Execution time of SOMTimeS, K-means, and TADPole  for the 112 archived UCR datasets on a (a) linear, and (b) log scaled axes.}
\label{SOMfig:scalability_executionTime}
\end{figure}

Finally, Figure~\ref{SOMfig:scalability_executionTime} shows how SOMTimeS execution time scales with the problem size ($\sum_{i=1}^{n}{|Q|_i}$, where $|Q|$ is the length of times series $Q$, and $n$ is the total number of time series in the dataset). It increases at a lower rate than TADPole and is similar to K-means (with DTW distance pruning). TADPole increases at the highest rate, consistent with its O($n^2$) complexity of DTW calculations, followed by K-means with a complexity of O($n \times k \times$ number of iterations), where $k$ is the number of clusters. SOMTimeS has complexity of O($n \times k \times e$), where $e$ is the number of epochs. K-means (at 10 iterations) is slightly faster per unit problem size than SOMTimeS (at 10 epochs) because it 1) has pruned slightly more DTW calculations, and 2) does not require weight updates. 
%

\section{Application to Serious Illness Conversations}\label{SOMsec:Case_Study}
We demonstrate the utility of a new artificial neural network -- SOMtimeS to visualize clustered output 
by applying it to healthcare communication setting using actual lexical data collected in the Palliative Care Communication Research Initiative (PCCRI) cohort study~\citep{Gramling2015}. The PCCRI is a multisite, epidemiological study that includes verbatim transcriptions of audio-recorded palliative care consultations involving 231 hospitalized people with advanced cancer, their families, and 54 palliative care clinicians.

\subsection{Need for scalability in health care communication science}

Understanding and improving healthcare communication requires 
a methodology that can
measure what actually happens when patients, families, and clinicians interact in large enough samples 
to represent 
diverse cultural, dialectical, decisional and clinical contexts~\citep{Tulsky2017}. 
Some features of inter-personal communication, such as tone or lexicon, will require frequent sampling over the course of conversation 
in order to reveal overarching patterns indicating types of interactions. Discovering patterns (i.e., clusters) 
of conversations with frequent sampling of 
features over conversation 
presents a need for scalable unsupervised machine learning methods. 
SOMTimeS is equipped to meet the need.

Our previous work suggests that 
conversational 
narrative analysis offers a clinically meaningful framework for understanding serious illness conversations~\citep{ROSS2020826,GRAMLING2021}, and others have demonstrated that unsupervised machine learning can identify “types of stories” using time-series analysis of lexicon~\citep{Reagan2016}. One core feature of conversational narrative, called \textit{temporal reference}, characterizes how participants organize 
their conversations about things that happened in the past, 
are happening now,  or 
may happen in the future~\citep{ROMAINE198387}. 
This motivates a study of how SOMtimeS can be useful 
to explore potential clusters of ``temporal reference story arcs''. Natural language processing methods can reasonably estimate  the shape or ``arc'' 
of temporal reference by categorizing verb tenses spoken during a conversation and describing the relative frequency of past/present/future referents over sequential deciles of total words spoken in the conversation (i.e., narrative time). In order to avoid sparse decile-level data 
in shorter conversations, we selected the 171 of 231 PCCRI clinical conversations as the basis for examining potential clustering.


\subsection{Data pre-processing: Verb tense as a time series}
%

\begin{figure}[!ht]
\centering
\begin{subfigure}[b]{0.45\textwidth}
\centering
\caption{}\label{SOMfig:all_conversations_raw}
\includegraphics[width=1\textwidth]{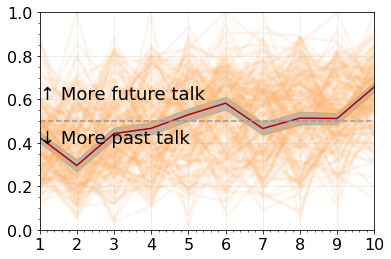}
\end{subfigure}
\begin{subfigure}[b]{0.45\textwidth}
\centering
\caption{}\label{SOMfig:all_conversations_smooth}
\includegraphics[width=1\textwidth]{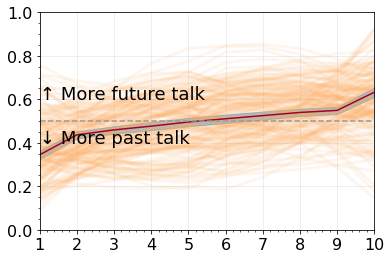}
\end{subfigure}
\caption{Temporal plot showing the (a) raw time series, and (b) smoothed time series for all conversations superimposed in brown; the red line represents the mean values, and the shaded region around the red line represents 95\% confidence interval.}\label{SOMfig:all_conversations}
\end{figure}

We used a temporal reference tagger~\citep{ROSS2020826} to assign temporal reference (past, present, or future) to verbs and verb modifiers in the verbatim transcripts. Specifically, the Natural Language Toolkit (NLTK; www.nltk.org) was 
used to classify each word in the transcripts 
into a part of speech (POS), and for any word classified as a verb, the preceding context is used to assign that verb (and any modifiers) 
to
a given temporal reference.
Then, each conversation was stratified into deciles of ``narrative time'' based on the total word count for each conversation, and a temporal reference (i.e., verb tense) time series was generated for each conversation as the proportion of all future tense verbs relative to the total number of past and future tense verbs. The 
vertical axis in Figure~\ref{SOMfig:all_conversations} 
represents the proportion of future vs. past talk (per decile), where any value above the threshold (dashed line = 0.5) represents more future talk. 
Each of the 171 
generated time series (see Figure~\ref{SOMfig:all_conversations_raw}) were then smoothed using a 2nd-order, 9-step Savitzky-Golay filter~\citep{Savitzky-Golay} (see Figure~\ref{SOMfig:all_conversations_smooth}). Savitzky-Golay filter works by fitting a polynomial over a moving window (2nd-order polynomial, over a 9-step window in this work) and replaces the data points with corresponding values of the fitted polynomial.
Smoothing reduces noise that may result from simplifying assumptions used in modeling the temporal reference time series (i.e., conversational story arcs). We then 
used SOMTimeS to cluster the resulting conversational story arcs. 

%

\subsection{Clustering verb tense time series}

\begin{figure}[!ht]
\centering
\includegraphics[width=0.4\textwidth]{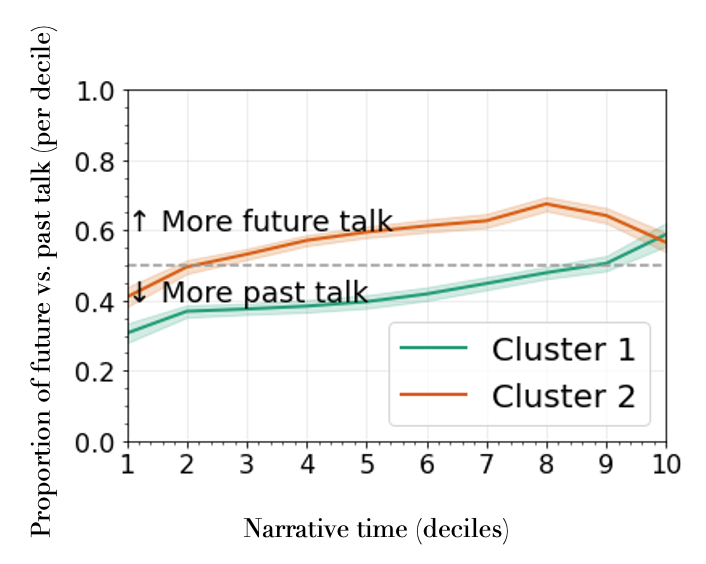}
\caption{Mean values of the proportion of future 
and past 
(i.e., verb tense) talks 
over the narrative time decile for cluster 1 (green) and cluster 2 (brown) with the shaded region representing 95\% confidence interval.}\label{SOMfig:application}
\end{figure}

In applying SOMTimeS to the conversational PCCRI data, we identified $k=2$ clusters with distinct temporal shapes (see Figure~\ref{SOMfig:application}). 
%
%
Both of the conversational arcs share a temporal narrative with more references to the past at the beginning of the conversation, and more references to the future as the conversation progresses. 
The 
proportions of future
talk 
and past talk are more similar 
at deciles 1 and 10 
than at deciles 2 
to 9. These conversational arcs are differentiated by the rate at which the narrative changes. Cluster 1 
does not enter the ``more future talk'' region until decile 9, while cluster 2 
does much earlier (decile 2). 
It was expected that the first and last deciles of the conversations 
would be more similar given the nature of introduction
at 
the
start and farewell
at the end
of a conversation. 

\begin{figure}[!ht]
\centering
\includegraphics[width=1\textwidth]{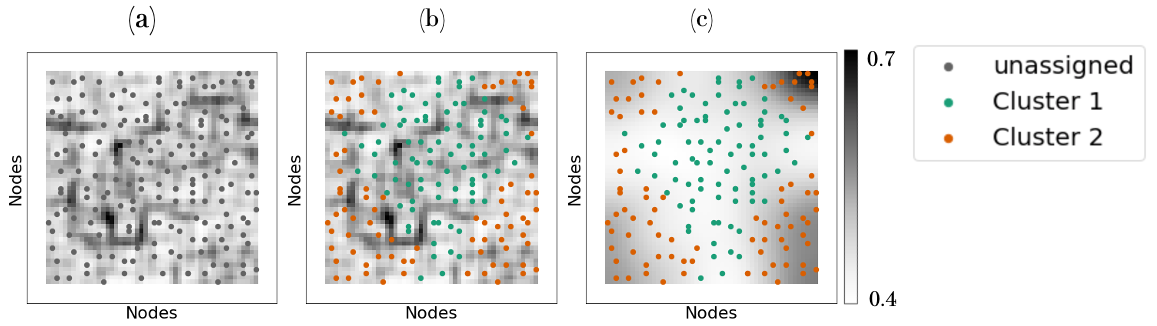}
\caption{Temporal reference time series data from 171 serious illness conversations self-organized on a 2-D map
(a) with U-matrix, (b) using spectral clustering, and 
(c) interpolated sum of the temporal reference time series
superimposed 
on the clustered map.}\label{SOMfig:Application_SOM}
\end{figure}

Now, let us illustrate how 
SOMTimeS identifies the number of clusters 
and 
visualizes the features 
that drive those clusters. When the U-matrix is superimposed on the 2-D SOM mesh (see Figure~\ref{SOMfig:Application_SOM}a), the observations appear to cluster into 2--3 groups based on visual inspection. Keeping the case study objectives in mind, and noting that the 2-D mesh is torodial, we color-coded the $k=2$ clusters on Figure~\ref{SOMfig:Application_SOM}b using spectral clustering. In Figure~see~\ref{SOMfig:Application_SOM}c, we superimpose and interpolate the sum of the proportion of future vs. past talk over all deciles in the time series (i.e., conversational arcs of Figure~\ref{SOMfig:application}) in the same 2-D space as the clustered times series.

\section{Discussion}\label{SOMsec:Discussions}

We present SOMTimeS as a clustering algorithm for time series that exploits the competitive learning of the Kohonen Self-Organizing Map, a pruning strategy and the distance bounds of DTW to improve execution time. SOMTimeS contrasts with other DTW-based clustering algorithms in both its ability to both reduce the dimensionality of, and visualize input features associated with clustering temporal data. We also implemented a similar DTW-distance pruning strategy in K-means for the first time to demonstrate performance gains achieved for what is likely the most popular clustering algorithm to date. In terms of accuracy, SOMTimeS has higher assessment indices compared to TADPole, and while the assessment indices are statistically similar with K-means, the additional functionality of the SOM comes with a higher computational cost. 

The benchmark experiments in this work are intended to put SOMTimeS in context with state-of-the-art time series clustering algorithms. Keeping the study objectives in mind, execution times are used to demonstrate scalability, and highlight the feasibility of analyzing large time series datasets using SOMTimeS. 
K-means is perhaps the most popular clustering algorithm and has been proven time and again to outperform state-of-the-art algorithms; however, because of its simplicity, it lacks the interpretability and visualization capabilities of SOMTimeS. TADPole on the other hand, is a state-of-the-art clustering algorithm that organizes data differently from SOMTimeS (and by extension K-means), as evident from the difference in ARI scores (see Figure~\ref{SOMfig:accuracy}a), and choice of centroids (i.e., density peaks; see Supplementary Material Section~\ref{SOMsec:TADPole}). For these reasons, the algorithms tested are not direct competitors of one another and each has advantages in their own right.

SOMTimeS learns (i.e., self-organizes) in an iterative manner such that as the number of SOM epochs increase, the execution time per epoch decreases (see Figure~\ref{SOMfig:overEpochs}b), making higher number of epochs (and thus, corresponding assessment indices) feasible. This reduction in time is also directly proportional to the number of calls to the DTW function at each epoch. The elbow point (at 6 for SOMTimeS with 100 epochs) indicates quick gains in pruning DTW calculations. This same gain is observed when the total number of epochs is set to 10 or 50 (see Supplementary Material Figure~\ref{SOMfig:overEpochs10}). SOMTimeS took 40 minutes to cluster the entire UCR archive using 10 epochs, and less than 300 minutes when the number of epochs was increased 10-fold. Similarly, the largest dataset in terms of problem size took 5 minutes to cluster using 10 epochs, and 35 minutes to cluster at 100 epochs. SOMTimeS demonstrates sub-linear scalability when it comes to increasing the number of epochs. 
The scalability, fast execution times, and the ease of saving the state (weights) of a SOM make SOMTimeS a potential candidate for an \emph{anytime} algorithm. It possesses the five most desirable properties of anytime algorithms~\citep{Zilberstein1995,Zhu2012}. 

\paragraph{Concluding Remarks:}  This paper presents a computationally efficient variant of the SOM that uses DTW as a distance measure and a DTW-distance pruning strategy. To put its performance in context, two other state-of-the-art algorithms have been presented, TADPole and K-means. All three use DTW-distance pruning, and each has their own strengths and weaknesses. Firstly, they organize data differently, and secondly, the SOM is often used for data visualization, dimensionality reduction, and feature selection. For these reasons a direct comparison of the advantages and disadvantages of each algorithm is not possible, and a speed-up in one does not make the other redundant. SOMTimeS has unique data visualization abilities that require the mesh size to be increased to a value higher then $k$. The pruning strategy presented in this work makes the latter feasible. However, if only classification (or hard clusters) are required, then $k-means$ is the faster and equally accurate clustering algorithm.


\section{Conclusion and Future Work}\label{SOMsec:Conclusion}
The explosion in volume of time series data has resulted in the availability of large unlabeled time datasets. In this work, we introduce \textbf{S}elf-\textbf{O}rganizing \textbf{M}aps for \textbf{t}ime \textbf{s}eries (\textbf{SOMTimeS}). SOMTimeS is a self-organizing map for clustering and classifying time series data that uses DTW as a distance measure of similarity between time series. To reduce run time and improve scalability, SOMTimeS prunes DTW calculations by using distance bounding during the SOM training phase. This pruning results in a computationally efficient and fast time series clustering algorithm that is linearly scalable with respect to increasing number of observations. SOMTimeS clustered 112 datasets from the UCR time series classification archive in under 5 hours with state-of-art accuracy. 
We also implemented a similar pruning strategy in K-means for the first time to demonstrate performance gains achieved for what is likely the most popular clustering algorithm to date.
We applied SOMTimeS to 171 conversations from the PCCRI dataset. The resulting clusters showed two fundamental shapes of conversational stories. 

To further improve computational efficiency and clustering accuracy, newer and state-of-the-art variations of SOMs may be used that leverage the same pruning strategy in this work. Improving computational time of DTW-based algorithms is an active area of research, and any improvement in computational speed of DTW can be incorporated in SOMTimeS for the unpruned DTW computations. Finally, SOMTimeS is a uni-variate time series clustering algorithm. To create a multivariate time series clustering algorithm, the pruning strategy will have to be revisited to accommodate the variations of DTW for multi-variate time series.
SOMTimeS is a fast and linearly scalable algorithm that recasts DTW as a computationally efficient distance measure for time series data clustering.

\section*{Acknowledgements}

This project was supported by the Richard Barrett Foundation and Gund Institute for Environment through a Gund Barrett Fellowship. Additional support was provided by the Vermont EPSCoR BREE Project (NSF Award OIA-1556770). We thank Dr. Patrick J. Clemins of Vermont EPSCoR, for providing support in using the EPSCoR Pascal high-performance computing server for the project. Computations were performed, in part, on the Vermont Advanced Computing Core. We also thank all the curators and administrators of the UCR archive. Data used to illustrate concepts in this paper arise from the Palliative Care Conversation Research Initiative (PCCRI). The PCCRI was funded by a Research Scholar Grant from the American Cancer Society (RSG PCSM124655; PI: Robert Gramling). 

\bibliographystyle{apalike}
\bibliography{main.bib} 

\section{Supplementary Material}\label{SOMsec:supplementaryMaterial}
\beginsupplement

\begin{figure}[!htb]
\centering

\begin{subfigure}[b]{0.3\textwidth}
\centering
\caption{}\label{SOMfig:problemSizea}
\includegraphics[width=1\textwidth]{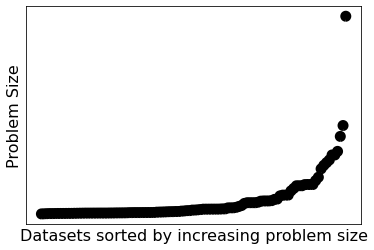}
\end{subfigure}
\quad%
\begin{subfigure}[b]{0.3\textwidth}
\centering
\caption{}\label{SOMfig:problemSizeb}
\includegraphics[width=1\textwidth]{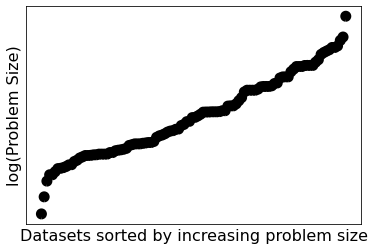}
\end{subfigure}
\caption{Distribution of all 128 datasets in the UCR archive in terms of (a) problem size and (b) natural log of problem size}.
\label{SOMfig:problemSize}
\end{figure}

\begin{figure}[!htb]
\centering

\begin{subfigure}[b]{0.4\textwidth}
\centering
\caption{DTW calculations over epochs}\label{SOMfig:dtwOverEpoch10}
\includegraphics[width=1\textwidth]{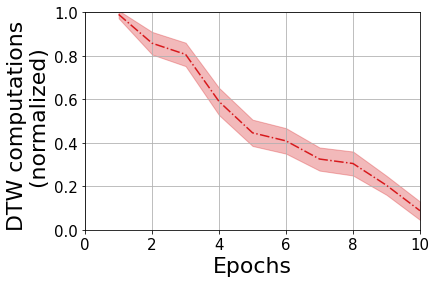}
\end{subfigure}
\quad%
\begin{subfigure}[b]{0.4\textwidth}
\centering
\caption{Execution time over epochs}\label{SOMfig:timeOverEpoch10}
\includegraphics[width=1\textwidth]{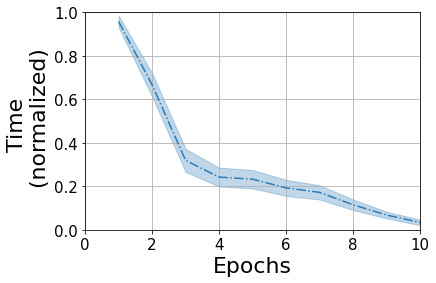}
\end{subfigure}
\caption{Change in pruning efficiency of SOMTimeS (10 epochs total) as reflected by the calls to DTW function, and execution time over epochs. The dotted line represents the mean value for all datasets after individually normalizing run for each dataset over all epochs. The shaded region corresponds to 95\% confidence interval around the mean.}
\label{SOMfig:overEpochs10}
\end{figure}

This supplementary material provides additional information on the following aspects of the study:

\begin{enumerate}

    \item Section~\ref{SOMsec:TADPole} describes the TADPole~\citep{Nurjahan2016} algorithm.
    \item Figure~\ref{SOMfig:problemSize} show datasets sorted by increasing problem size on arithmetic and log-log scale.
    \item Figure~\ref{SOMfig:timeOverEpoch10} shows the percentage of DTW computations by SOMTimeS, K-means, and TADPole for each of the 112 UCR archive datasets.
    \item Change in pruning efficiency of SOMTimeS (10 epochs total) as reflected by the calls to DTW function, and execution time over epochs. 
    
    
\end{enumerate}

\subsection{TADPole}\label{SOMsec:TADPole}

TADPole~\citep{Nurjahan2016} is a density based clustering method that uses Density Peaks~\citep{Rodriguez1492} as the clustering algorithm and DTW as the distance measure. 
The Density Peaks algorithm generates cluster centroids (called ``density peaks'') that are surrounded by neighboring data points that have lower local density and are relatively farther from data points with a higher local density~\citep{Rodriguez1492}.
The algorithm has two phases. It first finds centroids (density peaks), and then assigns data points to the closest centroid. The algorithm requires two input parameters: the number of clusters ($k$) and the local neighborhood distance $d$ (wherein the local density of a data point is calculated). In this work, when TADPole is used, $k$ is assumed to be known, and the value of $d$ is determined as the distance wherein the average number of neighbors is 1 to 2\% of the total number of observations in the dataset, following a rule of thumb proposed by the original authors~\citep{Rodriguez1492}. TADPole uses upper bound (Euclidean distance) and lower bound (LB\_Keogh) to prune unnecessary DTW calculations in the first phase to speed up the clustering. The algorithm has a complexity of O($n^2$) where n is the number of time series observations in the input.

\begin{figure}[!htb]
\centering

\includegraphics[width=0.8\textwidth]{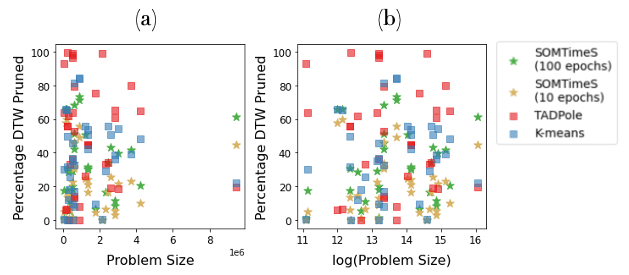}

\caption{The pruning effect of SOMTimeS (10 epochs shown in blown stars), SOMTimeS (100 epochs in green stars), K-means (blue squares) and TADPole (red squares) measured as the percentage of DTW calls pruned during the clustering of a dataset for varying problem size in (a) linear scale axis and (b) natural log axis.}\label{SOMfig:pruning_10_100}
\end{figure}

\end{document}